\documentclass{article}

\usepackage{spconf,amsmath,graphicx}

\usepackage[tight,footnotesize]{subfigure}
\usepackage{cite}
\usepackage{amsmath,amssymb,amsfonts}
\usepackage{algorithmic}
\usepackage{epstopdf}
\usepackage{float}
\usepackage{amsmath}
\usepackage{tabu}  
\usepackage[all]{xy}
\usepackage{graphics}
\usepackage{graphicx}
\usepackage{subfigure}
\graphicspath{{figures/}}
\usepackage{textcomp}
\usepackage{xcolor}
\usepackage{colortbl,booktabs}

\usepackage{algorithm}
\usepackage{algorithmic}
\usepackage{bm}
\usepackage{latexsym}
\usepackage{amsthm}
\usepackage{url}
\usepackage{amsfonts}
\usepackage{amssymb}
\usepackage{indentfirst}
\usepackage{float}

\newtheorem{definition}{Definition}
\newtheorem{problem}{Problem}

\usepackage{array}
\usepackage{boldline,multirow}
\usepackage{ragged2e}

\usepackage{amsmath}

\usepackage{scrextend}

\usepackage{algorithm}
\usepackage{algorithmic}
\usepackage{bm}
\usepackage{latexsym}
\usepackage{amsthm}
\usepackage{url}
\usepackage{amsfonts}
\usepackage{amssymb}
\usepackage{indentfirst}
\usepackage{float}

\title{AnomalyDAE: Dual autoencoder for anomaly detection on attributed networks}

\name{Haoyi Fan$^{1}$, Fengbin Zhang$^{1}$, Zuoyong Li$^{2}$\thanks{Corresponding authors: Fengbin Zhang, Zuoyong Li.}}
\address{$^{1}$School of Computer Science and Technology,\\
Harbin University of Science and Technology, Harbin 150080, China. \\ 
$^{2}$Fujian Provincial Key Laboratory of Information Processing and Intelligent Control,\\ 
Minjiang University, Fuzhou 350121, China.\\
\textit{\{isfanhy, zhangfengbin\}@hrbust.edu.cn, fzulzytdq@126.com}.
}

\begin{document}
%
\maketitle
\begin{abstract}
Anomaly detection on attributed networks aims at finding nodes whose patterns deviate significantly from the majority of reference nodes, which is pervasive in many applications such as network intrusion detection and social spammer detection. However, most existing methods neglect the complex cross-modality interactions between network structure and node attribute. In this paper, we propose a deep joint representation learning framework for anomaly detection through a dual autoencoder (AnomalyDAE), which captures the complex interactions between network structure and node attribute for high-quality embeddings. Specifically, AnomalyDAE consists of a structure autoencoder and an attribute autoencoder to learn both node embedding and attribute embedding jointly in latent space. Moreover, attention mechanism is employed in structure encoder to learn the importance between a node and its neighbors for an effective capturing of structure pattern, which is important to anomaly detection. Besides, by taking both the node embedding and attribute embedding as inputs of attribute decoder, the cross-modality interactions between network structure and node attribute are learned during the reconstruction of node attribute. Finally, anomalies can be detected by measuring the reconstruction errors of nodes from both the structure and attribute perspectives. Extensive experiments on real-world datasets demonstrate the effectiveness of the proposed method.
\end{abstract}
\section{Introduction}
\label{sec:intro}

Attributed networks \cite{li2019adaptive} are ubiquitous in the real world such as social networks\cite{liao2018attributed}, communication networks\cite{2019DPernesIJCNN}, and product co-purchase networks\cite{shi2018heterogeneous}, in which each node is associated with a rich set of attributes or characteristics, in addition to the raw network topology.

Anomaly detection on attribute networks aims at finding nodes whose patterns or behaviors significantly deviate from the reference nodes,  which has a broad impact on various domains such as network intrusion detection \cite{ding2012intrusion}, system fault diagnosis \cite{cheng2016ranking}, and social spammer detection \cite{fakhraei2015collective}. Recently, there is a growing interest in researches about anomaly detection on attributed networks. Some of them study the problem of community-level anomalies detection by comparing the current node with other reference nodes within the same community \cite{perozzi2016scalable} or measuring the quality of connected subgraphs \cite{perozzi2018discovering}. Some of them conduct anomaly analysis through subspace selection of node feature \cite{sanchez2013subspaces,perozzi2014focused}. While some recent residual analysis based methods attempt to find anomalies by assuming that anomalies cannot be approximated from other reference nodes \cite{li2017radar,peng2018anomalous}.

Although above mentioned algorithms had their fair share of success, these methods either suffer from severe computational overhead caused by shallow learning mechanisms and subspace selection, or neglect the complex interactions between nodes and attributes by only learning the representations for nodes\cite{ding2019deep}, while interactions between two different modality sources are of great importance for anomaly detection task to capture both structure and attribute induced anomalies. To alleviate the above-mentioned problems, in this paper, we propose a  deep joint representation learning framework for \textbf{anomaly} detection through a \textbf{d}ual \textbf{a}uto\textbf{e}ncoder (\textbf{AnomalyDAE}), which captures the complex interactions between the network structure and node attribute for high-quality embeddings. Different from \cite{ding2019deep}, which employs one single graph convolutional network (GCN) \cite{kipf2017semi} based encoder for node embedding, AnomalyDAE consists of a structure autoencoder and an attribute autoencoder to learn the latent representation of nodes and attributes jointly by reconstructing the original network topologies and node attributes respectively. Then, anomalies in the network are detected by measuring the reconstruction errors of nodes from both the structure and attribute perspectives.

In sum, the main contributions of this paper are as follows:

\begin{itemize}
\item We propose a  deep joint representation learning framework for anomaly detection on attributed networks via a dual autoencoder where the complex cross-modality interactions between the network structure and node attribute are captured, and the anomalies are measured from both the structure and attribute perspectives.
\item We conduct extensive experiments on multiple real-world datasets, and the results show that AnomalyDAE consistently outperforms state-of-the-art deep model significantly, with up to 22.32\% improvement in terms of the ROC AUC score. The source codes\footnote{\url{https://github.com/haoyfan/AnomalyDAE}} are publicly available. 
\end{itemize}

\section{Notations and Problem Statement}
\label{sec:notations}

In this section, we formally define the frequently-used notations and the studied problem. The notations used in this paper are summarized in Table \ref{tab:notations}.

\begin{definition}
\textbf{Attributed Network} $\boldsymbol{\mathcal{G}}=\{\boldsymbol{\mathcal{V}}, \boldsymbol{\mathcal{E}}, \textup{$\boldsymbol{\mathrm{X}}$} \}$ is defined as an undirected graph with M=$|\boldsymbol{\mathcal{V}}|$ nodes and $|\boldsymbol{\mathcal{E}}|$ edges, each of nodes is associated with a N=\textup{$|\boldsymbol{\mathrm{X}}|$} dimension attribute.
\end{definition}

\begin{problem}
Given an attributed network $\boldsymbol{\mathcal{G}}=\{\boldsymbol{\mathcal{V}}, \boldsymbol{\mathcal{E}}, \textup{$\boldsymbol{\mathrm{X}}$} \}$, our goal is to detect the nodes that are rare and differ significantly from the majority of the reference nodes in terms both the structure and attribute information of the nodes. More formally, we aim to learn a score function $f: \boldsymbol{\mathcal{V}}_i \mapsto y_i \in \mathbb{R}$, to classify sample $x_i$ based on the threshold $\lambda$:
\begin{equation}
\label{eq:def}
\begin{split}
y_{i}=\left\{\begin{matrix}
1, & if \ f(\boldsymbol{\mathcal{V}}_i)\geq \lambda, \\ 
0, & otherwise.
\end{matrix}\right.
\end{split}
\end{equation}
where $y_{i}$ denotes the label of sample $x_i$, with 0 being the normal class and 1 the anomalous class.
\end{problem}

\begin{table}
\caption{Notations.}
\label{tab:notations}
\centering
\begin{tabular}{cV{2}c}
\hlineB{2}
\textbf{Notation}    & \textbf{Description} \\
\hlineB{2}
\boldmath${\mathcal{G}}$    &  Attributed network.    \\
\hline
\boldmath${\mathcal{V}}$ & A set of nodes in network. \\
\hline
\boldmath${\mathcal{A}}$ & A set of node attributes in network. \\
\hline
\boldmath${\mathcal{E}}$ & A set of edges in network. \\
\hline
$M$ & The number of nodes. \\
\hline
$N$ & The dimension of attribute. \\
\hline
$D$ & The dimension of embedding. \\
\hline
 \textup{$\boldsymbol{\mathrm{A}}\in \mathbb{R}^{M \times M}$} & Adjacency matrix of a network. \\
 \hline
 \textup{$\boldsymbol{\mathrm{X}}\in \mathbb{R}^{M \times N}$} & Attribute matrix of all nodes. \\
 \hline
 \textup{$\boldsymbol{\mathrm{Z}^{\mathcal{V}}}\in \mathbb{R}^{M \times D}$} & Latent embedding of nodes. \\
 \hline    
\textup{$\boldsymbol{\mathrm{Z}^{\mathcal{A}}}\in \mathbb{R}^{N \times D}$} & Latent embedding of attributes. \\
\hlineB{2}    
\end{tabular}
\end{table}

\begin{figure*}
  \centering
  \includegraphics[width=7in]{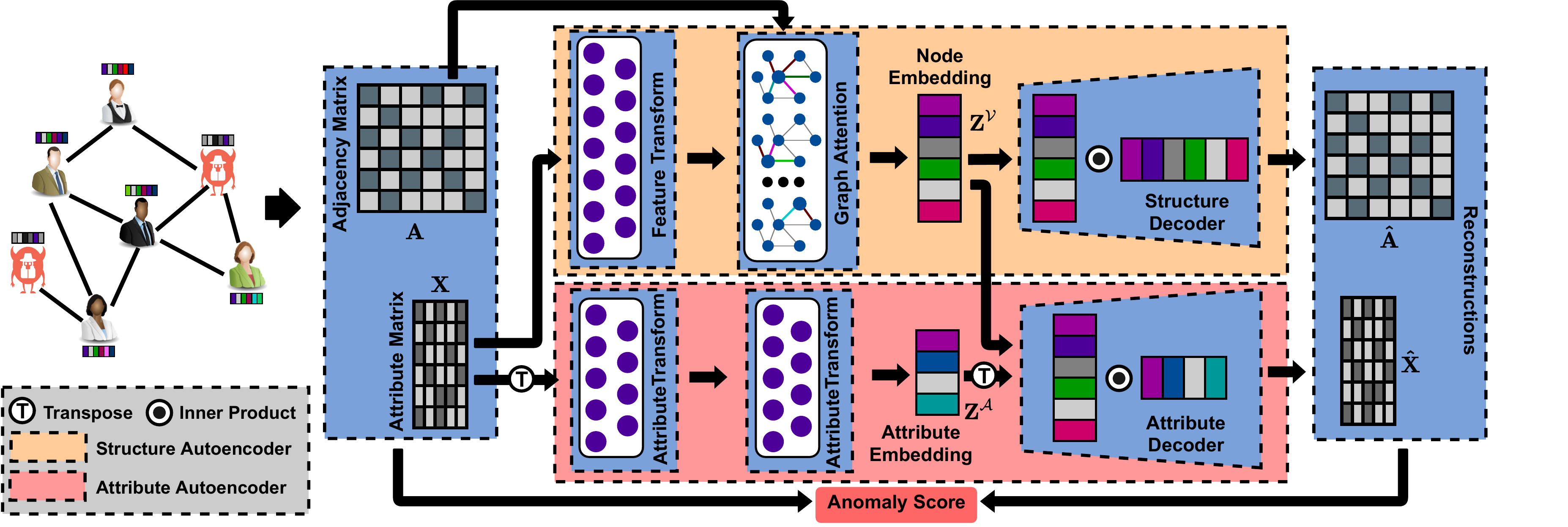}
 \caption{The framework of the proposed AnomalyDAE.}
\label{fig:framework}
\end{figure*}

\section{Method}
\label{sec:method}
In this section, we introduce the proposed AnomalyDAE in detail. As shown in Fig. \ref{fig:framework}, AnomalyDAE is an end-to-end joint representation learning framework that consists of a structure autoencoder for network structure reconstruction, and an attribute autoencoder for node attributes reconstruction. Take the learned node embedding from the structure encoder and the learned attribute embedding from the attribute encoder as inputs, the interactions between the network structure and the node attribute are jointly captured by both structure decoder and attribute decoder during the training. Finally, anomalies in the network can be measured by the reconstruction errors of network structure and node attribute.

\subsection{Structure Autoencoder}
\label{sec:structureAE}

In order to obtain sufficient representative high-level node features, structure encoder firstly transforms the original observed node attribute \textup{$\boldsymbol{\mathrm{X}}$} into the low-dimentional latent representation \textup{$\boldsymbol{\widetilde{\mathrm{Z}}^{\mathcal{V}}}$}, which is shown as follows:
\begin{equation}
\begin{split}
\label{eq:init_node_transform}
&\boldsymbol{\widetilde{\mathrm{Z}}^{\mathcal{V}}}=\sigma( \textup{$\boldsymbol{\mathrm{X}}$}\boldsymbol{\mathrm{W}^{\mathcal{V}_{(1)}}}+{\boldsymbol{\textbf{b}^{\mathcal{V}_{(1)}}}})
\end{split}
\end{equation}
where $\sigma(\bullet)$ is the activation function such as ReLU or Tanh, $\boldsymbol{\mathrm{W}^{\mathcal{V}_{(1)}}} \in \mathbb{R}^{D_{in}^{\mathcal{V}_{(1)}} \times D_{out}^{\mathcal{V}_{(1)}}} $ and ${\boldsymbol{\textbf{b}^{\mathcal{V}_{(1)}}}} \in \mathbb{R}^{D_{out}^{\mathcal{V}_{(1)}}} $ are the weight and bias learned by encoder, $D_{in}^{\mathcal{V}_{(1)}}$ and $D_{out}^{\mathcal{V}_{(1)}}$ are the dimensionalities of \textup{$\boldsymbol{\mathrm{X}}$} and $\boldsymbol{\widetilde{\mathrm{Z}}^{\mathcal{V}}}$, respectively.

Given the transformed node embedding $\boldsymbol{\widetilde{\mathrm{Z}}^{\mathcal{V}}}$, a graph attention layer \cite{velickovic2018graph} is then employed to aggregate the representation from neighbor nodes, by performing a shared attentional mechanism on the nodes:
\begin{equation}
\begin{split}
\label{eq:node_attn}
& e_{i,j} = attn(\boldsymbol{\widetilde{\mathrm{Z}}^{\mathcal{V}}}_{i}, \boldsymbol{\widetilde{\mathrm{Z}}^{\mathcal{V}}}_{j})=\sigma(\textbf{a}^{\mathrm{T}}\cdot [\boldsymbol{\mathrm{W}^{\mathcal{V}_{(2)}}} \boldsymbol{\widetilde{\mathrm{Z}}^{\mathcal{V}}}_{i}||\boldsymbol{\mathrm{W}^{\mathcal{V}_{(2)}}} \boldsymbol{\widetilde{\mathrm{Z}}^{\mathcal{V}}}_{j}])
\end{split}
\end{equation}
where $e_{i,j}$ is the importance weight of node $\boldsymbol{\mathcal{V}}_{i}$ to node $\boldsymbol{\mathcal{V}}_{j}$, $attn(\bullet)$ denotes the neural network parameterized by weights $\textbf{a} \in \mathbb{R}^{D}$ and $\boldsymbol{\mathrm{W}^{\mathcal{V}_{(2)}}} \in \mathbb{R}^{\frac{D}{2} \times D_{out}^{\mathcal{V}_{(1)}}}$ that shared by all nodes, $||$ denotes the concatenate operation. Then, the final importance weight $\gamma_{i,j}$ is normalized through the softmax function:
\begin{equation}
\begin{split}
\label{eq:attn_norm}
&{\gamma_{i,j}}=\frac{\mathrm{exp}(e_{i,j})}{\sum_{k \in \mathcal{N}_{i}}\mathrm{exp}(e_{i,k})}
\end{split}
\end{equation}
where $\mathcal{N}_{i}$ denotes the neighbors of node $\boldsymbol{\mathcal{V}}_{i}$, which is provided by adjacency matrix \textup{$\boldsymbol{\mathrm{A}}$}, and its final embedding $\boldsymbol{{\mathrm{Z}}^{\mathcal{V}}}_{i}$ can be obtained by weighted sum based on the learned importance weights as follows:
\begin{equation}
\begin{split}
\label{eq:final_node_embedding}
&\boldsymbol{{\mathrm{Z}}^{\mathcal{V}}}_{i}=\sum_{k \in \mathcal{N}_{i}}\gamma_{i,k}\cdot \boldsymbol{\widetilde{\mathrm{Z}}^{\mathcal{V}}}_{k}
\end{split}
\end{equation}

Finally, structure decoder takes the final node embeddings $\boldsymbol{\mathrm{Z}^{\mathcal{V}}}$ as inputs to decode them for reconstruction of the original network structure:
\begin{equation}
\label{eq:reconstruct_structure}
\begin{split}
\textup{$\boldsymbol{\hat{\mathrm{A}}}$}=Sigmoid(\boldsymbol{\mathrm{Z}^{\mathcal{V}}}(\boldsymbol{\mathrm{Z}^{\mathcal{V}}})^{\mathrm{T}})
\end{split}
\end{equation}
where $Sigmoid(\bullet)$ is the sigmoid activation function. Here, inner product between two node embeddings is performed to estimate the probability of a link between two nodes:
\begin{equation}
\label{eq:link_prob}
\begin{split}
p(\textup{$\boldsymbol{\hat{\mathrm{A}}}_{i,j}=1|\boldsymbol{\mathrm{Z}^{\mathcal{V}}}_{i},\boldsymbol{\mathrm{Z}^{\mathcal{V}}}_{j}$})=Sigmoid(\boldsymbol{\mathrm{Z}^{\mathcal{V}}}_{i}(\boldsymbol{\mathrm{Z}^{\mathcal{V}}}_{j})^{\mathrm{T}})
\end{split}
\end{equation}

\subsection{Attribute Autoencoder}
\label{sec:attributeAE}

In the attribute encoder, two non-linear feature transform layers are employed to map the observed attribute data to the latent attribute embedding \textup{$\boldsymbol{{\mathrm{Z}}^{\mathcal{A}}}$}, which can be formulated as follows:
\begin{equation}
\begin{split}
\label{eq:init_attr_transform}
&\boldsymbol{\widetilde{\mathrm{Z}}^{\mathcal{A}}}=\sigma((\textup{$\boldsymbol{\mathrm{X}}$})^{\mathrm{T}}\boldsymbol{\mathrm{W}^{\mathcal{A}_{(1)}}}+{\boldsymbol{\textbf{b}^{\mathcal{A}_{(1)}}}})
\end{split}
\end{equation}
\begin{equation}
\begin{split}
\label{eq:second_attr_transform}
&\boldsymbol{{\mathrm{Z}}^{\mathcal{A}}}=\boldsymbol{\widetilde{\mathrm{Z}}^{\mathcal{A}}}\boldsymbol{\mathrm{W}^{\mathcal{A}_{(2)}}}+{\boldsymbol{\textbf{b}^{\mathcal{A}_{(2)}}}}
\end{split}
\end{equation}
where $\boldsymbol{\mathrm{W}^{\mathcal{A}_{(1)}}} \in \mathbb{R}^{D_{in}^{\mathcal{A}_{(1)}} \times D_{out}^{\mathcal{A}_{(1)}}} $, ${\boldsymbol{\textbf{b}^{\mathcal{A}_{(1)}}}} \in \mathbb{R}^{D_{out}^{\mathcal{A}_{(1)}}}$, $\boldsymbol{\mathrm{W}^{\mathcal{A}_{(2)}}} \in \mathbb{R}^{ D_{out}^{\mathcal{A}_{(1)}} \times D} $ and ${\boldsymbol{\textbf{b}^{\mathcal{A}_{(2)}}}} \in \mathbb{R}^{D}$ are the weights and biases learned by two layers, $D_{in}^{\mathcal{A}_{(1)}}$, $D_{out}^{\mathcal{A}_{(1)}}$, and $D$ are the dimensionalities of \textup{$(\boldsymbol{\mathrm{X}})^{\mathrm{T}}$}, $\boldsymbol{\widetilde{\mathrm{Z}}^{\mathcal{A}}}$ and $\boldsymbol{{\mathrm{Z}}^{\mathcal{A}}}$, respectively. 

Finally, attribute decoder takes both the node embeddings $\boldsymbol{\mathrm{Z}^{\mathcal{V}}}$ learned by structure encoder, and the attribute embeddings $\boldsymbol{{\mathrm{Z}}^{\mathcal{A}}}$ as inputs for decoding of the original node attribute:
\begin{equation}
\label{eq:reconstruct_attribute}
\begin{split}
\textup{$\boldsymbol{\hat{\mathrm{X}}}$}=\boldsymbol{\mathrm{Z}^{\mathcal{V}}}(\boldsymbol{\mathrm{Z}^{\mathcal{A}}})^{\mathrm{T}}
\end{split}
\end{equation}
in which, the interactions between network structure and node attribute are jointly captured. Different from the structure decoder, in the attribute decoder, no activation function is utilized here for the arbitrary-valued attribute.

In AnomalyDAE, the computational complexity of structure autoencoder and attribute autoencoder are $O(MD+ED+M^{2})$ and $O(ND+NM)$ respectively, where $M$ is the number of nodes, $E$ is the number of edge, $N$ is the dimension of attribute, $D$ is the dimension of embedding.

\subsection{Loss function}
\label{sec:loss}
The training objective of AnomalyDAE is to minimize the reconstruction errors of both network structure and node attribute:
\begin{equation}
\label{eq:loss}
\begin{split}
\mathcal{L}_{rec}=\alpha||(\textup{$\boldsymbol{{\mathrm{A}}}$}-\textup{$\boldsymbol{\hat{\mathrm{A}}}$})\odot \boldsymbol{\theta}||^{2}_{F}+(1-\alpha)||(\textup{$\boldsymbol{{\mathrm{X}}}$}-\textup{$\boldsymbol{\hat{\mathrm{X}}}$})\odot \boldsymbol{\eta}||^{2}_{F}
\end{split}
\end{equation}
where $\alpha$ is the parameter which control the trade off between structure reconstruction and attribute reconstruction. $\odot$ is the Hadamard product, $\boldsymbol{\eta}$ and $\boldsymbol{\theta}$ are defined as follows:
\begin{equation}
\label{eq:loss2}
\begin{split}
\boldsymbol{\theta}_{i,j}=\left\{\begin{matrix}
1 & if \ \textup{$\boldsymbol{{\mathrm{A}}}$}_{i,j}=0, \\ 
\theta & otherwise.
\end{matrix}\right.
,
\boldsymbol{\eta}_{i,j}=\left\{\begin{matrix}
1 & if \ \textup{$\boldsymbol{{\mathrm{X}}}$}_{i,j}=0, \\ 
\eta & otherwise.
\end{matrix}\right.
\end{split}
\end{equation}
where $\eta >1$ and $\theta >1$, which are used to impose more penalty to the reconstruction error of the non-zero elements due to some missing edges or attributes in the real world.

\subsection{Anomaly Detection}
\label{sec:anomaly_detection}
Inspired by the motivation that the pattern of abnormal nodes deviate from the majority of other nodes in either structure or attribute, the anomaly score $\mathcal{S}_{\boldsymbol{\mathcal{V}}_{i}}$ of node $\boldsymbol{\mathcal{V}}_{i}$ is defined as the reconstruction error from both network structure and node attribute perspective:
\begin{equation}
\label{eq:anomaly_score}
\begin{split}
\mathcal{S}_{\boldsymbol{\mathcal{V}}_{i}}=\alpha||(\textup{$\boldsymbol{{\mathrm{A}}}_{i}$}-\textup{$\boldsymbol{\hat{\mathrm{A}}}_{i}$})\odot \boldsymbol{\theta}_{i}||^{2}_{F}+(1-\alpha)||(\textup{$\boldsymbol{{\mathrm{X}}}_{i}$}-\textup{$\boldsymbol{\hat{\mathrm{X}}}_{i}$})\odot \boldsymbol{\eta}_{i}||^{2}_{F}
\end{split}
\end{equation}
Based on the measured anomaly scores, the threshold $\lambda$ in Eq. \ref{eq:def} can be determined according to distribution of scores, e.g. the nodes of top-k scores are classified as anomalous nodes.


\begin{table}[t!]
\caption{Statistics of the used Real-World datasets.}
\label{tab:datasets}
\centering
\begin{tabular}{|c|c|c|c|c|}
\hline
Database	& \# $\mathcal{V}$	&\# $\mathcal{E}$ &\# $\mathcal{A}$ &\# Anomalies \\
\hline
\hline
BlogCatalog    &5,196  &171,743	 &8,189 &300	\\
Flickr    &7,575  &239,738	&12,047    &450	\\
ACM	&16,484  &71,980 	&8,337 	&600	\\

\hline	
\end{tabular}
\end{table}

\section{Experiments}
\label{sec:exp}

\subsection{Experimental Setup}
\label{subsec:datasets}
Three commonly used real-world datasets\cite{ding2019deep} are used in this paper to evaluate the proposed method, including BlogCatalog, Flickr, and ACM. The statistics of datasets are shown in Table \ref{tab:datasets}. In the experiment, we train AnomalyDAE with 100, 100, 80 iterations for BlogCatalog, Flickr, and ACM respectively. Adam\cite{kingma2015adam} algorithm is utilized for optimization with learning rate as 0.001. The embedding dimension  is set as 128 for all datasets. The parameters $(\alpha, \eta, \theta)$ are empirically set as (0.7, 5, 40), (0.9, 8, 90), (0.7, 3, 10) for BlogCatalog, Flickr, and ACM respectively.

\subsection{Result Analysis}
\label{subsec:result_analysis}

\subsubsection{Performance Evaluation}

We compare AnomalyDAE with the state-of-the-art methods including LOF \cite{breunig2000lof}, SCAN \cite{xu2007scan}, AMEN \cite{perozzi2016scalable}, Radar\cite{li2017radar}, Anomalous\cite{peng2018anomalous}, and Dominant\cite{ding2019deep}. The \textbf{AUC} scores (the \textbf{A}rea \textbf{U}nder a receiver operating characteristic \textbf{C}urve) for anomaly detection are reported in Table \ref{tab:auc_anomaly_detection}. 

The experimental results show that the proposed AnomalyDAE significantly outperforms all baselines in various datasets. The performance of AnomalyDAE is much higher than traditional anomaly detection methods, i.e., AnomalyDAE outperforms LOF by 48.66\%, SCAN by 70.54\%, and AMEN by 44.44\% respectively on AUC for BlogCatalog dataset, because of that LOF and SCAN consider only network structure or node attribute, and AMEN is designed for anomalous neighborhoods rather than node itself. Besides, for Flickr dataset, AnomalyDAE increases the AUC score by 24.36\% compared with Radar and 25.63\% compared with Anomalous, this is because that residual analysis or CUR decomposition\cite{mahoney2009cur} based methods are not only sensitive to network sparsity, but with limited  learning ability on large graph. Compared with more recent Dominant, AnomalyDAE achieves gains by 19.68\%, 22.32\%, 15.11\% on BlogCatalog, Flickr, and ACM datasets respectively. Although GCN based encoder in Dominant is capable of learning discriminative node embeddings by aggregating neighbor features, the single graph encoder cannot jointly capture the complex interactions between network structure and node attribute, while AnomalyDAE employs two separate encoders for jointly learning of node embedding and attribute embedding respectively, which considers the complex modality interactions of network structure and node attribute.

\begin{table}
\caption{AUC scores of all methods on three datasets.}
\label{tab:auc_anomaly_detection}
\centering
\begin{tabular}{V{2}cV{2}c|c|cV{2}cV{2}}

\hlineB{2}
Method    &{BlogCatalog}  &{Flickr} &{ACM}  \\
\hlineB{2}
LOF\cite{breunig2000lof} &49.15 &48.81 &47.38 \\

SCAN\cite{xu2007scan} &27.27 &26.86  &35.99 \\

AMEN\cite{perozzi2016scalable} &53.37 &60.47 &72.62 \\

Radar\cite{li2017radar} &71.04 &72.86 &69.36 \\

Anomalous\cite{peng2018anomalous} &72.81 &71.59 &71.85 \\

Dominant\cite{ding2019deep} &78.13  &74.9 &74.94 \\

\hline
AnomalyDAE &\textbf{97.81} &\textbf{97.22}  &\textbf{90.05} \\
\hlineB{2}
\end{tabular}
\end{table}

\begin{figure}
\begin{minipage}[b]{.48\linewidth}
  \centering
  \centerline{\includegraphics[height=3.2cm]{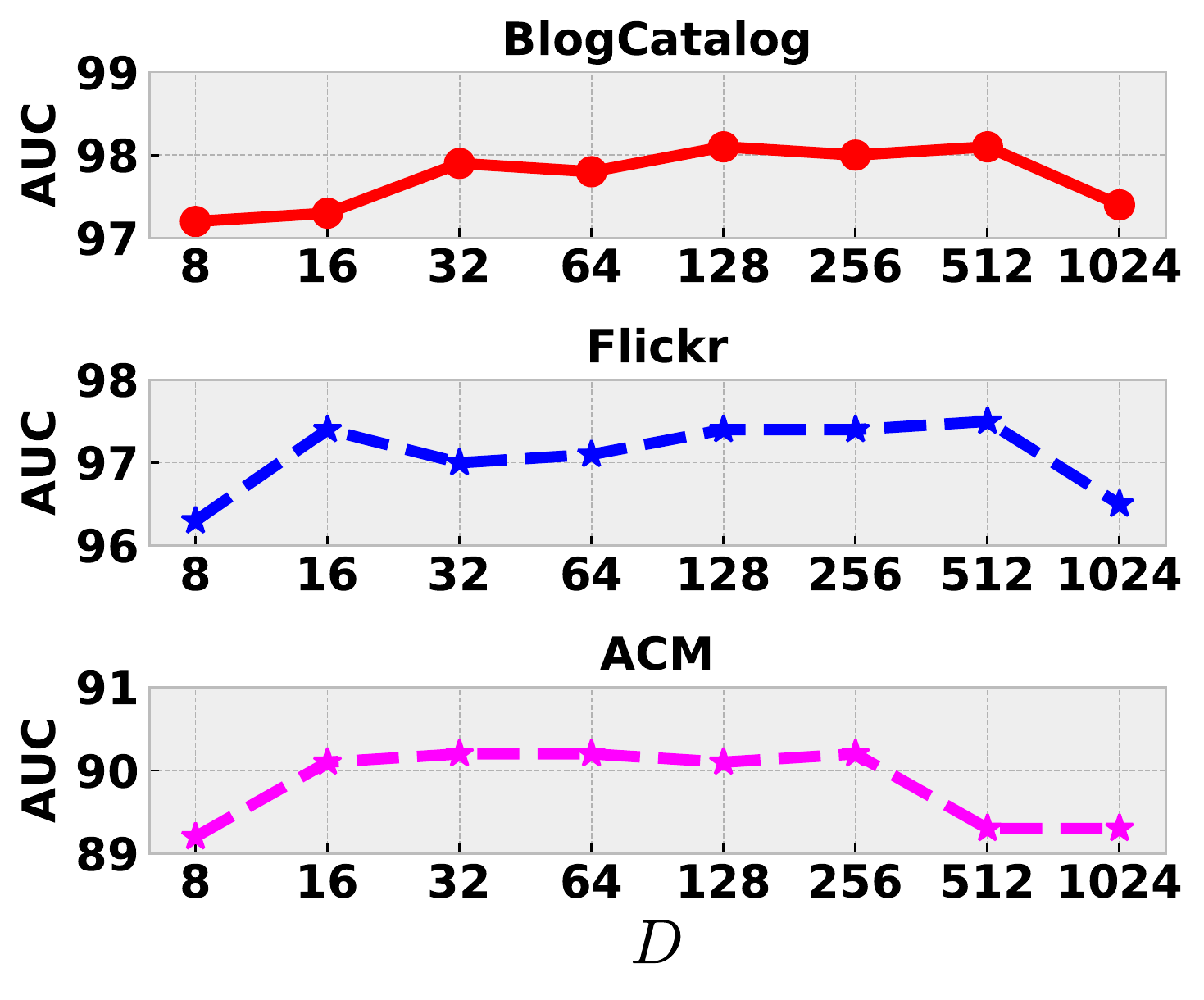}}
  \centerline{(a) Embedding dimension}\medskip
\end{minipage}
\hfill
\begin{minipage}[b]{0.48\linewidth}
  \centering
  \centerline{\includegraphics[height=3.2cm]{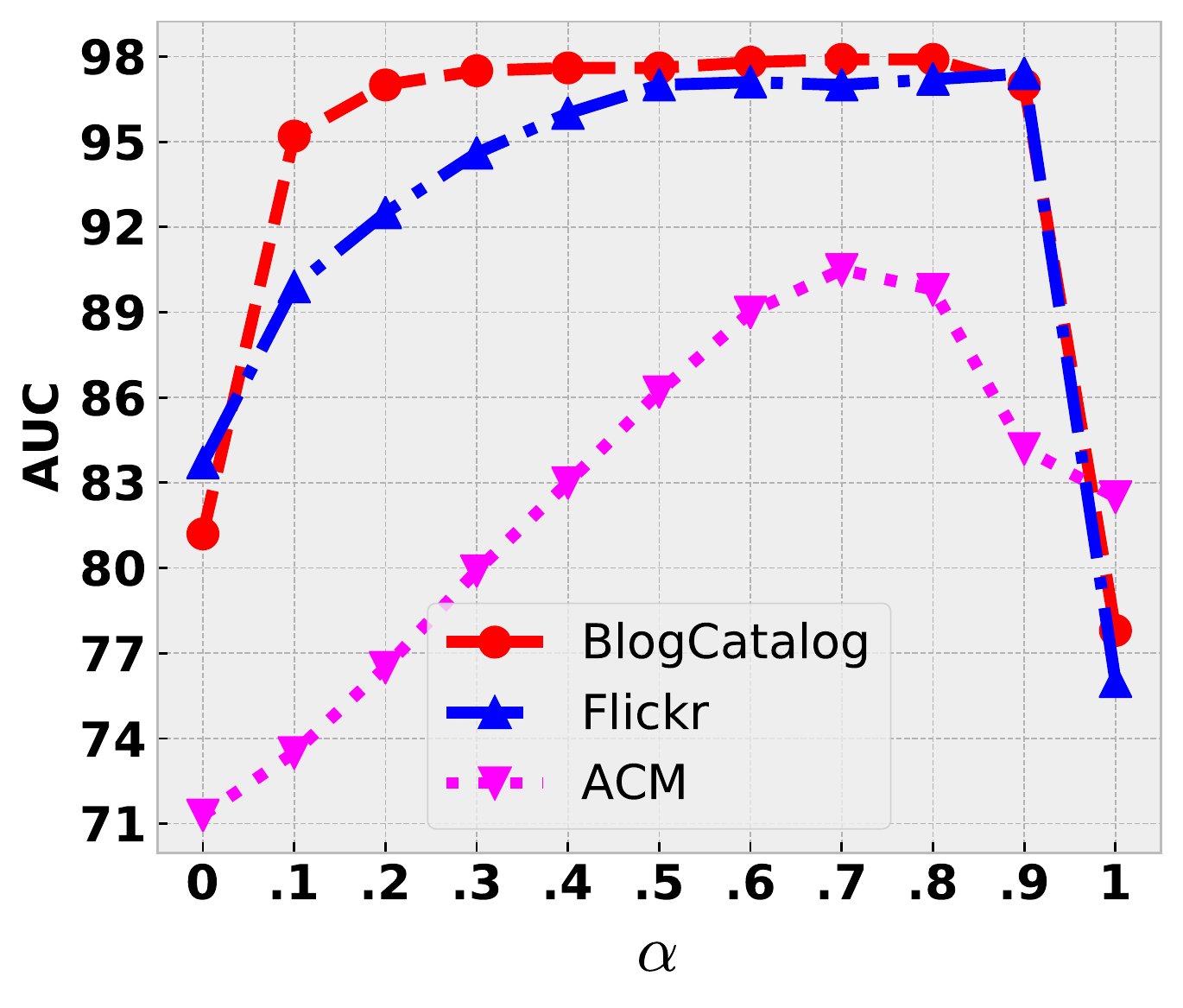}}
  \centerline{(b) Parameter $\alpha$ }\medskip
\end{minipage}
\caption{Parameter sensitivity for anomaly detection.}
\label{fig:parameter_sentivity}
\end{figure}

\subsubsection{Parameter Sensitivity}
In this section, we investigate the parameter sensitivity of different numbers of the embedding dimension $D$ and the trade-off parameter $\alpha$ for anomaly detection. The experiment results are shown in Fig. \ref{fig:parameter_sentivity}. We can see that a relative high dimension such as 128 or 256 facilitates high performance because higher dimensional embeddings are capable of encoding more information. However, the dimension with too low value or too high value would degrade the performance because of the weak modeling ability or suffering from overfitting. In terms of $\alpha$, considering only attribute reconstruction ($\alpha$=0) or structure reconstruction ($\alpha$=1) would results in poor performance, which demonstrates the importance of the interactions between the network structure and node attribute on attributed network for anomaly detection.

\section{Conclusion}
\label{sec:conclusion}
In this paper, we study the problem of anomaly detection on attributed network by considering the complex modality interaction between network structure and node attribute. To cope with this problem, we propose a  deep joint representation learning framework for anomaly detection via a dual autoencoder. By introducing two separate autoencoders for jointly learning of node embedding and attribute embedding, AnomalyDAE performs better than current state-of-the-art methods on multiple real-world datasets.

\bibliographystyle{IEEEbib}
\bibliography{strings,refs}

\end{document}